\documentclass{article}
\usepackage[preprint]{neurips_2026}
\usepackage[T1]{fontenc}
\usepackage[utf8]{inputenc}
\usepackage{amsmath,amssymb,amsthm,booktabs,graphicx,microtype,tabularx}

\usepackage{flafter}
\usepackage{placeins}
\usepackage{xcolor,url}
\usepackage[colorlinks=true,linkcolor=blue,citecolor=blue,urlcolor=blue]{hyperref}
\hypersetup{pdftitle={Technical Report on RUNE: Beyond Alignment in MEG-to-Text Decoding},pdfauthor={Yiqian Yang and Yiqun Duan}}
\title{Technical Report on RUNE:\\[0.2em]Beyond Alignment in MEG-to-Text Decoding}
\author{Yiqian Yang \qquad Yiqun Duan\\Uploading Inc\\
\texttt{yiqian@uploading.tech} \qquad \texttt{yiqunduan@uploading.tech}}
\date{November 23, 2026}
\begin{document}
\maketitle
\begin{abstract}
Non-invasive brain-to-text seeks to recover language from brain activity without surgery. A central challenge is turning learned brain--language representations into accurate word predictions. We present RUNE, a decoding system that combines corrected training and evaluation, brain--text alignment, a class-balanced word readout, and target-word language-model fusion. On LibriBrain100's word-ranking task, RUNE reaches \textbf{64.12\%} mean Top-10 balanced accuracy, improving on published MEG-XL by \textbf{16.92 percentage points} across four corpora, with gains on each corpus. Controlled comparisons show that both implementation corrections and a learned readout contribute to this improvement. Candidate-level analysis reveals that some words receive informative scores yet consistently lose to competing candidates, while an identical-feature comparison shows that learning the classification rule improves decoding without changing the representation. These findings identify the decoding rule as a consequential part of non-invasive brain-to-text and show how system improvements can make better use of aligned neural features.
\end{abstract}

\section{Introduction}
\label{sec:introduction}

\begin{figure}[t]
  \centering
  \includegraphics[width=\linewidth]{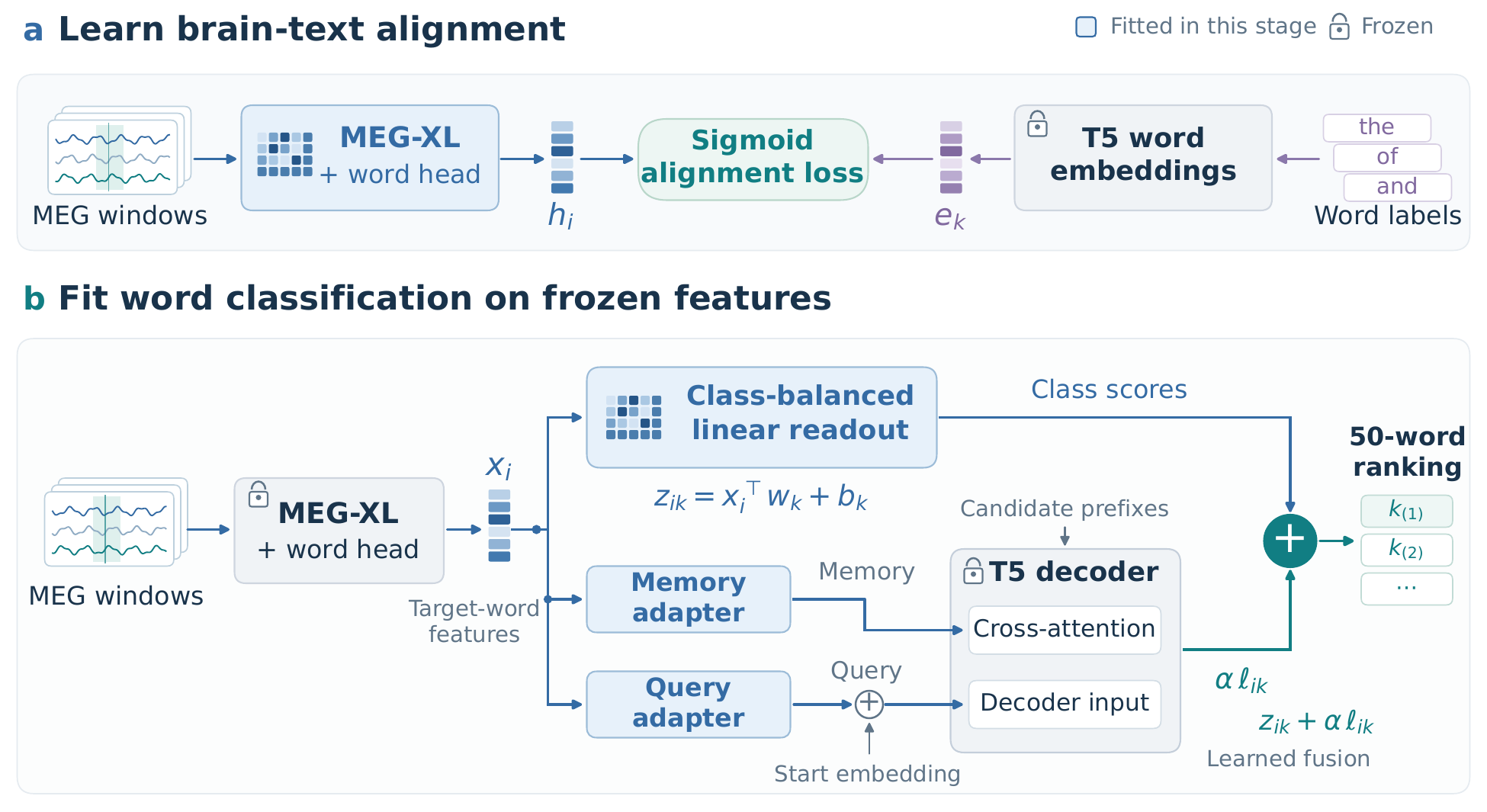}
  \caption{RUNE: from brain--text alignment to word prediction. \textbf{(a)} The encoder and word head learn alignment against fixed T5 embeddings using the corrected training and evaluation pipeline. \textbf{(b)} A class-balanced linear readout learns to rank candidate words from the frozen features. Two adapters map the target event to the memory and start-query inputs of a frozen T5 decoder, whose word scores are fused with the readout scores. Shaded blue modules are trained in the indicated stage; locks identify frozen modules. Signal traces are schematic.}
  \label{fig:method}
\end{figure}

Recovering language from non-invasive brain recordings could help restore communication for people with severe paralysis. Progress toward this goal requires both learning useful neural representations and converting them into accurate language predictions. MEG provides millisecond-resolution measurements of brain activity without surgery, and LibriBrain100 \citep{libribrain100} makes more than one hundred hours of speech-listening recordings available for studying this problem. Its word-decoding benchmark offers a concrete test: given a neural event at a known word onset, rank the corresponding word among fifty candidates.

The MEG-XL baseline \citep{megxl} approaches this task by aligning neural features with frozen T5 word embeddings \citep{t5} and ranking words by their similarity to those embeddings. This connects representation learning and prediction through a single scoring rule. Yet a score that distinguishes a word's own events from other events can still remain below competing words' scores. We observe exactly this pattern in a retrained MEG-XL model: some words have informative scores but never enter the top ten. The observation raises a broader question for decoding systems: how can we turn the information available after alignment into more accurate word predictions?

RUNE addresses this question across the decoding pipeline (Figure~\ref{fig:method}). We first correct defects in the released training and evaluation code so that the alignment model is trained and evaluated as intended. We then retain the learned representation and fit a class-balanced classifier that learns how candidate words compete. Finally, a frozen T5 decoder, conditioned on the target event through learned adapters, supplies complementary word scores. This design preserves brain--text alignment while allowing the prediction rule to adapt to the word-decoding task.

We evaluate the complete system, isolate implementation and readout improvements through controlled comparisons, and diagnose which words gain or lose correct predictions. Our contributions are:
\begin{itemize}
  \item \textbf{A stronger complete MEG decoding system.} RUNE combines implementation repairs, aligned neural features, learned classification and target-word T5 fusion. It improves performance across all four benchmark corpora, with controlled experiments measuring the gains from specific stages (Section~\ref{sec:decomposition}).
  \item \textbf{Evidence that similarity scoring leaves useful information untapped.} Learning a class-balanced readout on identical frozen features improves every corpus on both validation and test. This comparison establishes that better word predictions are possible without changing the representation, while retaining a linear scoring rule (Section~\ref{sec:readout}).
  \item \textbf{A diagnostic distinction between separation and retrieval.} Word-level analysis identifies informative scores that fail in the shared candidate ranking. Comparing readouts shows how recovering missed words trades off against predictions elsewhere in the vocabulary, giving a way to diagnose decoding failures beyond aggregate accuracy (Sections~\ref{sec:mechanism}--\ref{sec:capacity}).
\end{itemize}
Code and reproduction: \url{https://github.com/OpenUploading/RUNE}.

\section{Related Work}
\label{sec:related}

\paragraph{Speech decoding by alignment.}
Contrastive alignment enables identification of heard speech by connecting non-invasive brain recordings to pretrained speech representations \citep{defossez2023}. MEG-XL \citep{megxl} learns long-context MEG representations, and its LibriBrain100 baseline \citep{libribrain100} adapts them to word classification through alignment with T5 embeddings \citep{t5} using a SigLIP-style loss \citep{siglip}. We retain this representation-learning approach and examine the complete path from its implementation to the rule used for word prediction.

\paragraph{Similarity scoring and fitted classifiers.}
Linear probes test how well frozen representations support classification \citep{alain2016probes}. CLIP compares fitted linear classifiers with predictions based on text embeddings \citep{clip}, and long-tailed recognition separates representation learning from classifier fitting \citep{decoupling}. These precedents motivate our controlled comparison. The MEG encoder already receives word supervision during alignment, so our question is whether fitting the downstream decision rule improves predictions from that supervised representation. Word-level diagnostics then examine what changes in the resulting ranking.

\paragraph{Hubness and class priors.}
In high-dimensional nearest-neighbour retrieval, some targets become hubs that are returned much more often than others \citep{hubness}. This imbalance has motivated neighbourhood-based score corrections for zero-shot retrieval \citep{dinu2015hubness,csls,invertedsoftmax}; logit adjustment addresses class-prior effects in long-tailed classification \citep{logitadjustment}. The uneven candidate frequencies we observe on LibriBrain100 motivate testing both families of correction. They provide comparisons for determining how much can be recovered by adjusting similarity scores while retaining the text embeddings as classifier weights.

\section{Method}
\label{sec:method}

\paragraph{Representation.}
Each labelled word occurrence defines an event $i$, with an MEG window spanning $-0.5$ to $+2.5$\,s relative to its onset. Following the original implementation, the encoder jointly processes groups of up to 50 consecutive, overlapping windows. A pooling and projection head produces a 1,024-dimensional vector $h_i$ for each event (Appendix~\ref{app:selection}). Unless stated otherwise, readouts use its frozen, normalized representation $x_i=h_i/\lVert h_i\rVert_2$. We denote the event's word label by $y_i$ and the normalized frozen T5 embedding of candidate $k$ by $\hat e_k=e_k/\lVert e_k\rVert_2$. The $K=50$ candidate words are the classification labels.

\paragraph{Similarity scoring as a linear classifier.}
Similarity scoring ranks candidate $k$ by $\cos(h_i,e_k)=x_i^{\top}\hat e_k$. A linear readout ranks it by
\begin{equation}
  z_{ik}=x_i^{\top}w_k+b_k .
  \label{eq:linear}
\end{equation}
Similarity scoring fixes $w_k=\hat e_k$ and $b_k=0$. The fitted readout learns these weights and biases from labelled events while keeping $x_i$ unchanged. Both rules therefore score the same representation with a linear function; the intervention changes how its directions are combined and how scores are offset across candidate words. This provides a direct test of the decision rule's contribution to decoding.

\paragraph{Fitted readouts.}
The \emph{cross-entropy readout} is class-weighted multinomial logistic regression on $x_i$. With $n_k$ the number of training events of class $k$ and $a_i=1/n_{y_i}$, we fit $W$ and $b$ in Equation~\ref{eq:linear} by minimizing
\begin{equation}
  \mathcal{L}(W,b)=
  -\frac{\sum_i a_i\log\operatorname{softmax}(x_i^{\top}W+b)_{y_i}}{\sum_i a_i}
  +\frac{\lambda}{2}\lVert W\rVert_F^2 ,
  \label{eq:competitive}
\end{equation}
The softmax trains each word's score relative to all other candidates, and inverse-frequency weights give each class equal total weight during fitting. This emphasizes word-level coverage rather than the raw frequency of correct events. The bias is unpenalized. We fit on training labels and choose the regularization strength on pooled validation data (Appendix~\ref{app:selection}).

To test the choice of objective, we fit a second linear readout using \emph{squared loss} against one-hot word labels. It uses the same features, class weights and unpenalized bias, with a ridge penalty selected on validation data. This comparison holds the representation and classifier form fixed while changing how prediction errors are penalized.

\paragraph{T5 branch.}
The T5 branch tests whether a pretrained language decoder contributes information beyond the fitted classifier. Two trainable adapters map neural features to a memory sequence for cross-attention and a query added to the start-token embedding of a frozen T5-large decoder. We compare memory from all events in the input group (\emph{group memory}) with memory from the target event alone (\emph{target-only memory}). Target-only memory retains the target's contextualized feature from joint encoding of the group. The decoder normalizes next-token scores over allowed continuations in a candidate prefix tree and sums log-probabilities along each word's path to obtain $\ell_{ik}$ (Appendix~\ref{app:selection}). The system ranks words by $z_{ik}+\alpha\ell_{ik}$, combining classifier and decoder scores with $\alpha\ge0$ fitted on training data. The complete RUNE system uses target-only memory.

\section{Experimental Setup}
\label{sec:setup}

\paragraph{Data.}
We use the public LibriBrain100 training release for Subjects 0--32 and the official Subject~0 validation and test partitions for Sherlock, TIMIT, MOCHA-TIMIT and TheMoth.\footnote{The dataset is distributed as \texttt{pnpl/LibriBrain2}: \url{https://huggingface.co/datasets/pnpl/LibriBrain2}.} Encoder alignment uses 15,613 groups comprising 780,650 events; Subject~0 contributes 95.87\% of these events. Readout fitting uses the 314,371 training events whose labels belong to the fifty-word candidate vocabulary; events outside that vocabulary contribute to alignment but are excluded from readout fitting. Validation and test contain 3,149 and 3,172 fixed-vocabulary events, respectively, covering all fifty word classes. Appendix~\ref{app:reproduction} gives the corpus-level counts.

\paragraph{Metric and baseline.}
We report the equal-weight four-corpus mean of Top-10 balanced accuracy. Within each corpus, we compute the fraction of correct top-ten predictions separately for each word class and average over the classes present. A uniform random ranking has expected accuracy of 20\%. For diagnostics, a candidate is \emph{nominated} whenever it enters an event's top ten; its nomination rate is the fraction of pooled test events where this occurs, irrespective of the true word. We compare with published joint-training MEG-XL scores \citep{libribrain100}, whose four-corpus mean is 47.20\% (Table~\ref{tab:headline}), using the same official partitions, candidate list and metric. All differences are computed before rounding. A separate matched implementation control, described in Appendix~\ref{app:control}, measures the effect of the repairs.

\paragraph{Backbone.}
We initialize the encoder from the official MEG-XL V2 pretrained weights, freeze the signal tokenizer, and train the encoder, word head and loss parameters. The original alignment objective uses other words in each group as negatives and masks repeated words. We repair two defects in the released implementation: evaluation left dropout active, and optimization and checkpoint saving omitted the loss temperature and bias. All fitted readouts use the epoch-2 model selected on validation data after two epochs of training (checkpoint~A). Archived diagnostic scores use the epoch-3 model selected after continuing the run to six epochs (checkpoint~B).

\paragraph{Selection and test use.}
Within each variant, validation data determine the hyperparameters and training epoch, and training labels are used to fit the readout (Appendix~\ref{app:selection}). Table~\ref{tab:all} reports all 20 development variants; Section~\ref{sec:limitations} describes how test feedback entered system selection.

\paragraph{Statistical analysis.} We estimate sampling uncertainty by resampling sentences within each corpus, using 10,000 cluster-bootstrap replicates \citep{bootstrap}. Every method is evaluated on the same resampled sentences, giving paired score differences. We report 95\% basic bootstrap intervals; Appendix~\ref{app:reproduction} defines these intervals and provides the percentile alternatives. The identical-feature comparison uses checkpoint~A, whose similarity score is 60.21\%. The saved event-level comparisons and diagnostics use fitted readouts on checkpoint~A and similarity scoring on checkpoint~B, whose mean is 59.98\%.

\section{Results}
\label{sec:results}

\subsection{Main comparison}
\label{sec:decomposition}

\begin{table}[t]
  \centering
  \small
  \caption{Top-10 balanced accuracy (\%) on the official LibriBrain100 Subject~0 test partitions, fifty candidates. The published values are those for training on all subjects jointly, given to one decimal as in the source. RUNE improves the four-corpus mean by \textbf{16.92 percentage points}. Table~\ref{tab:system} compares our readouts on identical features.}
  \label{tab:headline}
  \begin{tabular}{lrrrrr}
\toprule
Method & Sherlock & TIMIT & MOCHA-TIMIT & TheMoth & Mean \\
\midrule
Published MEG-XL baseline & 49.2 & 43.1 & 50.1 & 46.4 & 47.20 \\
Full system (ours) & \textbf{70.78} & \textbf{58.71} & \textbf{71.17} & \textbf{55.83} & \textbf{64.12} \\
\bottomrule
\end{tabular}

\end{table}

\begin{table}[t]
  \centering
  \small
  \caption{Readout ablation: Top-10 balanced accuracy (\%) of five readouts of the same frozen features (checkpoint~A). $\Delta$ is the change in the four-corpus test mean relative to similarity scoring.}
  \label{tab:system}
  \resizebox{\linewidth}{!}{\begin{tabular}{lrrrrrrr}
\toprule
 & \multicolumn{6}{c}{Test} & Validation \\
\cmidrule(lr){2-7}\cmidrule(lr){8-8}
Readout on checkpoint A & Sherlock & TIMIT & MOCHA-TIMIT & TheMoth & Mean & $\Delta$ & Mean \\
\midrule
Similarity scoring & 64.86 & 58.02 & 68.56 & 49.40 & 60.21 & --- & 58.88 \\
Squared-loss readout & 70.20 & 55.19 & 64.44 & 55.00 & 61.21 & $+1.00$ & 62.29 \\
Cross-entropy readout & 70.58 & 58.55 & 70.95 & 55.83 & 63.98 & $+3.77$ & 64.03 \\
Cross-entropy readout + T5 branch, group memory & 70.48 & 58.73 & 70.95 & 55.83 & 64.00 & $+3.79$ & 64.23 \\
Cross-entropy readout + T5 branch, target-only memory (full system) & 70.78 & 58.71 & 71.17 & 55.83 & 64.12 & $+3.91$ & 64.22 \\
\bottomrule
\end{tabular}
}
\end{table}

The complete RUNE system reaches \textbf{64.12\%} four-corpus mean Top-10 balanced accuracy, a \textbf{16.92-point improvement} over published MEG-XL (Table~\ref{tab:headline}). The improvement holds separately on Sherlock, TIMIT, MOCHA-TIMIT and TheMoth.

\begin{figure}[t]
  \centering
  \includegraphics[width=\linewidth]{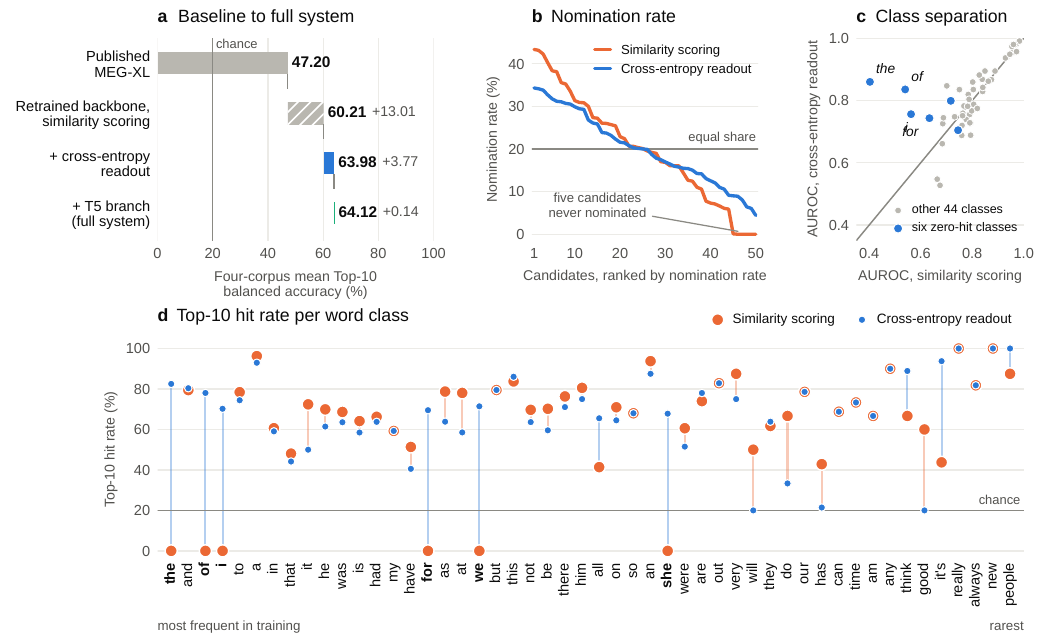}
  \caption{System improvement and word-level ranking behavior. \textbf{(a)} The 16.92-point gain over published MEG-XL; the hatched backbone step includes pretraining, public-data configuration, code repairs and retraining. \textbf{(b)} Candidate nomination rates on pooled test events, sorted independently for each readout, reveal words that similarity scoring never retrieves. \textbf{(c)} Per-class one-versus-rest AUROC compares separation of each word's events; points above the diagonal favor cross-entropy. \textbf{(d)} Per-class Top-10 hit rates, ordered by training frequency, show recovery of the six zero-hit classes marked in bold. At coincident values, the blue marker lies inside the orange marker. Panels (b)--(d) use archived scores specified in Section~\ref{sec:setup}; Table~\ref{tab:perclass} gives the values and class counts.}
  \label{fig:evidence}
\end{figure}

The implementation control and readout ablations identify improvements at successive stages of the system (Figure~\ref{fig:evidence}a). With our initialization, public data and two-epoch training configuration, the unmodified official code reaches 53.04\%. Correcting the two implementation defects raises similarity-scoring accuracy to 60.21\%, a matched gain of 7.17 points (Appendix~\ref{app:control}). Keeping these backbone features fixed, the learned readout adds 3.77 points and target-word T5 fusion adds a further 0.14 (Table~\ref{tab:system}).

These measured increments account for 11.08 points of the system improvement. The other 5.84 points separate the published result from our unmodified-code control and reflect the difference between those experimental configurations.

\subsection{Readout comparison}
\label{sec:readout}

\begin{table}[t]
  \centering
  \small
  \caption{Top-$k$ balanced accuracy (four-corpus test mean, \%) and paired Top-10 differences with 95\% intervals. Paired differences are against checkpoint~B (Section~\ref{sec:setup}); the same-feature differences are in Table~\ref{tab:system}.}
  \label{tab:readout}
  \begin{tabular}{lrrrl}
\toprule
Readout & Top-1 & Top-5 & Top-10 & Paired Top-10 difference from row 1 \\
\midrule
Similarity scoring (checkpoint B) & 16.54 & 43.52 & 59.98 & --- \\
Squared-loss readout & 12.30 & 42.36 & 61.21 & $+1.23$ $[-1.66,\,+3.18]$ \\
Cross-entropy readout & 16.32 & 46.95 & 63.98 & $+4.00$ $[+1.02,\,+5.72]$ \\
\midrule
Chance & 2.00 & 10.00 & 20.00 & \\
\bottomrule
\end{tabular}

\end{table}

On identical frozen features, the cross-entropy readout raises the four-corpus test mean from 60.21\% to 63.98\% (Table~\ref{tab:system}). Every corpus improves on both validation and test. Because the encoder outputs are unchanged, this comparison attributes the gain to learning the word-classification rule and demonstrates useful information left untapped by similarity scoring.

The archived event scores support a paired comparison with checkpoint-B similarity scoring: the cross-entropy readout improves the mean by $4.00$ points, with a 95\% interval of $[+1.02,+5.72]$ (Table~\ref{tab:readout}).

Fitting a classifier is only part of the improvement: the choice of loss also matters in this comparison. With features and class weights fixed, cross-entropy exceeds squared loss by 2.77 test points (paired 95\% interval $[+0.83,+4.43]$) and scores higher on validation. Squared loss improves over similarity scoring in two corpora and falls below it in the other two (Table~\ref{tab:system}). The objective that explicitly compares candidate scores produces the more consistent improvement.

The ranking gains are concentrated beyond the first position. In the archived-score comparison, cross-entropy changes Top-1 balanced accuracy by $-0.21$ points $[-2.17,+1.91]$ and Top-5 by $+3.44$ $[+0.52,+5.07]$ relative to checkpoint-B similarity scoring. Squared loss lowers Top-1 by 4.23 points. Thus, the improved Top-10 result reflects better inclusion of the true word among leading candidates, with little change in the cross-entropy readout's Top-1 score.

\subsection{Per-class analysis}
\label{sec:mechanism}

\begin{table}[t]
  \centering
  \small
  \caption{Diagnosing candidate competition through score corrections and fitted readouts. Six corrections retain the text embeddings as classifier weights; an in-sample reference and two learned readouts complete the comparison. Mean is four-corpus Top-10 balanced accuracy (\%). Hit rates average the pooled per-class Top-10 rates separately for the six zero-hit words and the remaining 44. Nomination rates show the minimum and maximum across candidates. Cross-fitted corrections are fitted on half of the test sentences and evaluated on the other half, then averaged over random splits; fitting and selection details appear in Appendix~\ref{app:reproduction}.}
  \label{tab:diagnostics}
  \resizebox{\linewidth}{!}{\begin{tabular}{llrrrrr}
\toprule
 & & & \multicolumn{2}{c}{Mean hit rate (\%)} & \multicolumn{2}{c}{Nomination rate (\%)} \\
\cmidrule(lr){4-5}\cmidrule(lr){6-7}
Readout & Labels used & Mean & six classes & other 44 & min & max \\
\midrule
Similarity scoring (checkpoint B) &  & 59.98 & 0.0 & 71.6 & 0.00 & 43.38 \\
\quad + per-candidate standardization & none & 57.13 & 21.7 & 64.3 & 13.43 & 31.05 \\
\quad + CSLS & none; setting chosen on test & 60.09 & 2.3 & 70.8 & 0.16 & 46.34 \\
\quad + inverted softmax & none; setting chosen on test & 59.51 & 65.1 & 60.6 & 2.81 & 60.97 \\
\quad + log-prior adjustment & training prior; setting chosen on test & 60.56 & 0.0 & 71.0 & 0.00 & 56.24 \\
\quad + fitted temperature and bias & test, cross-fitted & 56.57 & 16.6 & 64.2 & --- & --- \\
\quad + metric-maximizing offsets & test, cross-fitted & 56.43 & 11.3 & 64.2 & --- & --- \\
\quad + metric-maximizing offsets, in-sample reference & test, fitted and scored on the same events & 64.31 & 9.7 & 72.4 & 0.00 & 56.05 \\
\midrule
Squared-loss readout & training set & 61.21 & 69.9 & 64.4 & 6.43 & 35.37 \\
Cross-entropy readout & training set & 63.98 & 73.2 & 67.7 & 4.41 & 34.33 \\
\bottomrule
\end{tabular}
}
\end{table}

Similarity scoring consistently misses six words that account for 943 of the 3,172 test events. Five---\emph{the}, \emph{of}, \emph{we}, \emph{for} and \emph{she}---never enter the top ten; \emph{i} enters three times, but never when it is correct. At the other extreme, \emph{were} is nominated on 43.38\% of all events (Figure~\ref{fig:evidence}b, Table~\ref{tab:diagnostics}). The missed words include the first, third and fourth most frequent training candidates, while other common words such as \emph{and}, \emph{to} and \emph{a} have high hit rates. These \emph{zero-hit classes} therefore reveal a systematic ranking imbalance that affects frequently observed words as well as less common ones.

One-versus-rest AUROC distinguishes missed predictions from uninformative scores by testing whether a candidate's score is higher on its own events than on other events. Similarity scores for \emph{the} and \emph{of} show weak separation, but \emph{we} and \emph{she} reach AUROCs of 0.75 and 0.72 despite never entering the top ten (Figure~\ref{fig:evidence}c). Their scores contain information about the correct class, yet competing words outrank them. A zero hit rate can thus hide two different problems: weak class separation and a scoring rule that fails to turn available separation into retrieval.

The fitted cross-entropy readout nominates every candidate. In the archived-score comparison, the six zero-hit classes reach hit rates of 68--83\% (Figure~\ref{fig:evidence}d) and account for 734 of the 837 newly correct events. Most recovered hits therefore come from words that similarity scoring consistently missed.

In this diagnostic comparison, the six recovered classes contribute $+8.57$ points to the four-corpus mean, while changes in the remaining 44 contribute $-4.57$ (Appendix~\ref{app:reproduction}). The net improvement therefore comes from recovering persistently missed words at a smaller cost elsewhere in the vocabulary. Table~\ref{tab:perclass} reports the per-word separation and hit rates, showing which words benefit and which lose correct predictions.

Recovering the zero-hit words alone does not explain why cross-entropy is the better readout. Squared loss also nominates every word and gives the six classes a mean hit rate of 69.9\%, with class separation comparable to cross-entropy (Table~\ref{tab:diagnostics}). The difference lies in predictions elsewhere: on identical features, cross-entropy correctly ranks 167 events that squared loss misses, while the reverse occurs for 84 events. Its contribution to balanced accuracy improves for 29 classes and declines for 11. A useful readout must therefore recover neglected words while preserving correct rankings across the rest of the vocabulary.

Score corrections provide a complementary test of this ranking problem. Six evaluated corrections retain the text embeddings as weights and score below the learned readout (Table~\ref{tab:diagnostics}). The zero-initialized in-sample offset reference reaches a similar aggregate score while still missing several zero-hit words; another initialization gives all six positive hit rates (Appendix~\ref{app:reproduction}). These results show why aggregate accuracy and word-level recovery should be examined together. The appendix gives the corrections' fitting rules, including the use of test labels for diagnostic fitting and selection.

\subsection{Alternative readouts and the T5 branch}
\label{sec:capacity}

\begin{table}[t]
  \centering
  \small
  \caption{Nine alternative readouts on checkpoint~A (four-corpus test mean, \%). The first seven add a learned term to the frozen scores of the cross-entropy readout; the last two are fitted with a squared loss on other features. $\Delta$ is the difference from the linear readout with the same loss, with its paired 95\% interval where per-event scores were saved.}
  \label{tab:ablations}
  \begin{tabular}{llrrl}
\toprule
Alternative readout & Loss & Mean & $\Delta$ & Paired 95\% interval \\
\midrule
Cross-entropy readout & cross-entropy & 63.98 & & \\
Squared-loss readout & squared & 61.21 & & \\
\midrule
Nonlinear correction & cross-entropy & 63.75 & $-0.23$ & $[-0.71,\,+0.33]$ \\
Nonlinear correction with LayerNorm & cross-entropy & 63.36 & $-0.62$ & $[-1.37,\,+0.14]$ \\
Recurrent context correction & cross-entropy & 63.57 & $-0.41$ & $[-2.30,\,+1.13]$ \\
Cross-layer correction, equal weights & cross-entropy & 63.89 & $-0.09$ & --- \\
Cross-layer correction, gated & cross-entropy & 63.87 & $-0.11$ & --- \\
Corpus-specific bias & cross-entropy & 63.30 & $-0.67$ & --- \\
Corpus-specific linear correction & cross-entropy & 62.51 & $-1.47$ & --- \\
Intermediate-layer readout & squared & 61.32 & $+0.11$ & --- \\
Kernel readout & squared & 63.57 & $+2.36$ & $[+0.64,\,+3.92]$ \\
\bottomrule
\end{tabular}

\end{table}

After establishing the linear readout, we test whether additional transformations improve its predictions. Seven alternatives add nonlinear, contextual, cross-layer or corpus-specific corrections to its frozen scores (Table~\ref{tab:ablations}). None has a higher test mean than linear cross-entropy. For the three with saved event-level predictions, paired intervals include zero and their upper bounds are below 1.2 points. The gated cross-layer correction improves validation accuracy but scores lower on test. Within these tested designs, additional readout capacity does not provide a consistent improvement.

The remaining two alternatives change the input features of the squared-loss readout. Using intermediate word-head features after the GELU gives similar accuracy to using the final representation. A kernel feature map improves the squared-loss result, but its test mean remains below linear cross-entropy. In these comparisons, changing the fitting objective is more effective than either feature extension.

The T5 branch tests whether a pretrained decoder adds useful predictions to the learned readout. Target-only memory substantially outperforms group memory when each branch predicts words alone, on both validation and test, but their fused results are close (Table~\ref{tab:t5}, Appendix~\ref{app:selection}). Target-only fusion adds 0.14 points to the linear readout, bringing RUNE to 64.12\%. Thus, the large standalone advantage translates into only a small advantage after fusion: the relevant system-level comparison is the improvement beyond the existing classifier.

\section{Discussion}
\label{sec:discussion}

The system gain reflects improvements in both learning and using aligned features. The implementation control establishes a gain from correcting training and evaluation; the identical-feature comparison shows that the resulting representation supports better predictions than its original similarity scores achieve. Evaluating an aligned encoder through a single scoring rule can therefore understate its value for word decoding.

Word-level diagnostics help locate this gap. Informative scores with no nominations identify candidates whose cross-candidate ranking deserves investigation before concluding that their representation lacks useful information. The identical-feature readout comparison separately demonstrates that predictions can improve while the encoder remains fixed.

Both fitted objectives recover the six zero-hit words, but cross-entropy better balances recovered and lost predictions. The tested extensions do not surpass this linear readout, and T5 adds a small gain after fusion. These results favor establishing a class-balanced decision rule first, then judging extensions by their net contribution to the complete system.

\section{Limitations}
\label{sec:limitations}

The study uses one training seed and test-informed development: the full system was selected by test mean among 20 variants, and some score corrections were fitted or tuned on test labels. Bootstrap intervals quantify test-sampling uncertainty, excluding retraining and between-session variation. The benchmark evaluates Subject~0 with known word onsets and fifty candidates; this subject contributes about 96\% of training events. Overlapping windows expose timing cues \citep{simpleb2t}, some TIMIT windows share samples across partitions, and possible training exposure to MOCHA-TIMIT test sentences remains unverified. Generalization to unseen participants and open-vocabulary decoding therefore requires separate evaluation. The published-baseline comparison also includes differences in pretraining, public-data coverage and training configuration, while the original paper leaves the exact checkpoint and launch configuration unspecified. Our matched control measures the two repairs jointly with one seeded export; the residual gap to the published baseline remains unattributed. The 3.77-point readout gain holds features fixed, but has no computed paired interval. The $+4.00$ interval and six-class diagnostic use different checkpoints and do not isolate the readout's effect or establish the cause of the zero-hit pattern. Finally, the fitted classifier requires labelled examples, and the comparisons omit training-fitted offsets, standalone nonlinear cross-entropy classifiers and encoder fine-tuning. In-sample offsets match the fitted readout's aggregate score, so the design recommendations apply to the evaluated readouts and T5 configurations.

\section{Conclusion}
\label{sec:conclusion}

RUNE improves non-invasive word decoding by addressing the full path from brain--text alignment to candidate ranking. The complete system reaches 64.12\% mean Top-10 balanced accuracy on LibriBrain100, exceeding published MEG-XL by 16.92 percentage points with gains across all four corpora. Controlled comparisons establish improvements from correcting training and evaluation and from fitting a class-balanced readout to the same aligned features. Word-level analysis shows why the decoding rule matters: a candidate can distinguish its own events yet consistently lose to competing words. In this setting, similarity-based retrieval underestimates the word-decoding performance attainable from the aligned representation. Comparing it with a learned decision rule gives a more informative assessment of what these neural features support.

\clearpage
\begingroup
\small
\bibliographystyle{plainnat}
\bibliography{references}
\endgroup
\clearpage
\appendix

\section{Complete Results and Implementation Details}
\label{app:selection}

Table~\ref{tab:all} lists the 20 variants evaluated on the official Subject~0 test partitions. Identifiers in parentheses are those used in the repository.

\begin{table}[h]
  \centering
  \small
  \caption{All 20 variants: Top-10 balanced accuracy (\%) per corpus and four-corpus test mean. The first seven rows are backbones scored by similarity.}
  \label{tab:all}
  \resizebox{\linewidth}{!}{\begin{tabular}{lrrrrr}
\toprule
Variant & Sherlock & TIMIT & MOCHA-TIMIT & TheMoth & Mean \\
\midrule
Similarity scoring, checkpoint A & 64.86 & 58.02 & 68.56 & 49.40 & 60.21 \\
Similarity scoring, checkpoint B & 65.94 & 54.72 & 66.53 & 52.71 & 59.98 \\
Conditional temporal pooling & 66.09 & 55.69 & 69.78 & 49.16 & 60.18 \\
Static temporal pooling (epoch 2) & 64.30 & 55.73 & 69.91 & 50.18 & 60.03 \\
Static temporal pooling (epoch 3) & 65.70 & 57.19 & 67.57 & 51.35 & 60.45 \\
Spatial bottleneck (epoch 2) & 63.65 & 53.91 & 66.53 & 49.39 & 58.37 \\
Spatial bottleneck (epoch 5) & 63.12 & 55.76 & 65.58 & 51.24 & 58.93 \\
Squared-loss readout (E2) & 70.20 & 55.19 & 64.44 & 55.00 & 61.21 \\
Cross-entropy readout (E3) & 70.58 & 58.55 & 70.95 & 55.83 & 63.98 \\
Nonlinear correction (E6) & 70.82 & 58.63 & 70.13 & 55.41 & 63.75 \\
Intermediate-layer readout, squared loss (E7) & 69.85 & 55.64 & 64.85 & 54.92 & 61.32 \\
Kernel readout, squared loss (E8) & 71.71 & 59.04 & 67.39 & 56.13 & 63.57 \\
Recurrent context correction (E9) & 67.63 & 58.75 & 69.49 & 58.42 & 63.57 \\
Nonlinear correction with LayerNorm (E11) & 68.99 & 58.16 & 70.37 & 55.91 & 63.36 \\
Cross-layer correction, equal weights (E12) & 66.69 & 58.03 & 73.72 & 57.11 & 63.89 \\
Cross-layer correction, gated (E13) & 68.35 & 57.01 & 73.06 & 57.05 & 63.87 \\
T5 branch, group memory, fused (E14) & 70.48 & 58.73 & 70.95 & 55.83 & 64.00 \\
T5 branch, target-only memory, fused: full system (E15) & 70.78 & 58.71 & 71.17 & 55.83 & 64.12 \\
Corpus-specific bias (E16) & 70.92 & 59.23 & 67.32 & 55.75 & 63.30 \\
Corpus-specific linear correction (E17) & 71.04 & 59.26 & 64.46 & 55.28 & 62.51 \\
\bottomrule
\end{tabular}
}
\end{table}

\begin{table}[h]
  \centering
  \small
  \caption{Training and selection record. Pooled validation is balanced accuracy after combining the four validation corpora; corpus-mean validation is the four-corpus mean.}
  \label{tab:training}
  \begin{tabularx}{\linewidth}{lXX}
\toprule
Variant & Training or fitting & Selection \\
\midrule
Similarity scoring, checkpoint A & 2 epochs; 31,226 updates & Pooled validation; epoch 2 \\
Squared-loss readout (E2) & 3 closed-form ridge fits & Pooled validation \\
Cross-entropy readout (E3) & 3 L-BFGS fits; 1,856/577/108 iterations & Pooled validation; $\lambda=10^{-6}$ \\
Nonlinear correction (E6) & 4 configurations; 30 epochs each & Pooled validation; epoch 10 \\
Intermediate-layer readout (E7) & 2 layers; 3 penalties each & Corpus-mean validation; layer after the GELU \\
Kernel readout (E8) & 3 kernel widths; 3 penalties each & Pooled validation \\
Recurrent context correction (E9) & 2 widths; 20 epochs each & Corpus-mean validation; epoch 8 \\
Nonlinear correction with LayerNorm (E11) & 30 epochs; 18,450 updates & Corpus-mean validation; epoch 10 \\
Cross-layer correction, equal weights (E12) & 30 epochs; 18,450 updates & Corpus-mean validation; epoch 16 \\
Cross-layer correction, gated (E13) & 30 epochs; 18,450 updates & Corpus-mean validation; epoch 12 \\
T5 branch, group memory (E14) & 30 epochs; 36,870 updates & Corpus-mean validation; epoch 10 \\
T5 branch, target-only memory (E15) & 30 epochs; 36,870 updates & Corpus-mean validation; epoch 20 \\
Corpus-specific bias (E16) & One L-BFGS fit; 227 iterations & No selection (single converged fit) \\
Corpus-specific linear correction (E17) & One L-BFGS fit; 1,097 iterations & No selection (single converged fit) \\
\bottomrule
\end{tabularx}

\end{table}

\begin{table}[h]
  \centering
  \small
  \caption{T5 branches (four-corpus test mean, \%), scored alone and fused with the cross-entropy readout (63.98\%).}
  \label{tab:t5}
  \begin{tabular}{lrrr}
\toprule
T5 branch & Alone & Fused & Fusion weight $\alpha$ \\
\midrule
Group memory (E14) & 20.34 & 64.00 & 0.00319 \\
Target-only memory (E15) & 61.77 & 64.12 & 0.0674 \\
\bottomrule
\end{tabular}

\end{table}

\paragraph{Backbone.}
The backbone receives 306 MEG channels sampled at 50\,Hz and uses the original BioCodec signal tokenizer, which remains frozen. We initialize the medium-size MEG-XL V2 encoder from weights pretrained for 113,000 steps on five datasets. The word head averages the encoder output over time and projects it through a hidden layer of width 2,048 to a 1,024-dimensional output, with dropout 0.1. Feature-wise linear modulation conditions this head on 64-dimensional embeddings of 44 source, corpus and participant identifiers. Training uses AdamW with learning rates $10^{-5}$ for the encoder and $10^{-3}$ for the word head and loss parameters, weight decay $10^{-4}$ except on the loss temperature and bias, and gradient clipping at 1.0. We retain the original ReduceLROnPlateau schedule (factor 0.5, patience 5), use seed~42, and process one 50-word group per update; evaluation groups may be shorter. The alignment targets are frozen T5-large layer-12 word vectors. In the sigmoid loss, pairs with coincident targets (cosine similarity at least 0.999) receive zero weight, masking repetitions of the same word within a group.

\paragraph{Checkpoints.}
Checkpoint~A was selected by pooled validation among the first two epochs (31,226 updates); all downstream readouts use these epoch-2 features. The run was then continued to six epochs (93,678 updates), of which epoch~3 was selected on validation data (checkpoint~B). Relative to checkpoint~A, checkpoint-B similarity scores change by $+1.07$, $-3.30$, $-2.02$ and $+3.31$ test points on Sherlock, TIMIT, MOCHA-TIMIT and TheMoth. On validation, similarity scoring on checkpoint~A reaches 65.17\%, 53.06\%, 63.31\% and 54.00\% on Sherlock, TIMIT, MOCHA-TIMIT and TheMoth (mean 58.88\%), and the cross-entropy readout 71.18\%, 62.80\%, 63.71\% and 58.44\% (mean 64.03\%).

\paragraph{Backbone variants.}
Three backbone variants test alternatives to the original feature aggregation and are evaluated by similarity scoring (Table~\ref{tab:all}). Conditional temporal pooling learns input-dependent weights over time; static temporal pooling learns fixed temporal weights for each channel; a spatial bottleneck projects the 306 input channels onto 32 learned components. The table records the selected epoch within the first two training epochs and, for continued runs, the later selected epoch.

\paragraph{Fitted readouts.}
We fit the cross-entropy readout from zero initialization on all 314,371 fixed-vocabulary training events using full-batch L-BFGS \citep{lbfgs}. For $\lambda\in\{10^{-6},10^{-5},10^{-4}\}$, pooled validation accuracy is 64.38\%, 64.17\% and 62.17\%, respectively, selecting $\lambda=10^{-6}$. The corresponding four-corpus validation means are 64.03\%, 64.69\% and 63.59\%; selection uses the pooled criterion recorded in Table~\ref{tab:training}. Optimization uses double precision, a strong-Wolfe line search, and two phases with gradient tolerances $10^{-7}$ and $10^{-8}$. We accept a fit when its largest absolute gradient component is below $10^{-6}$. In increasing order of $\lambda$, the fits terminate after 1,856, 577 and 108 iterations with maximum gradient magnitudes of $3.4\times10^{-8}$, $1.3\times10^{-7}$ and $4.1\times10^{-7}$. All three fits take about nine minutes on one GPU. The squared-loss readout is solved in closed form.

\paragraph{Alternative readouts.}
Seven alternatives add corrections to the frozen cross-entropy scores. E6 uses a zero-initialized nonlinear network on normalized features, and E11 adds LayerNorm after its first linear layer. E9 supplies context through a two-layer bidirectional GRU over the word group. E12 and E13 project final and intermediate word-head features to 512 dimensions and combine them with equal weights or learned channelwise gates, respectively. E16 adds corpus-specific class biases, and E17 adds corpus-specific linear terms. Two further alternatives fit squared-loss readouts on different features: E7 uses the 2,048-dimensional activations after the word head's GELU, and E8 uses a 1,024-centre Nystr\"om approximation to an RBF kernel. The gated cross-layer correction reaches a validation mean of 66.39\% versus 64.03\% for linear cross-entropy, but their test means are 63.87\% and 63.98\%, respectively.

\paragraph{T5 branch.}
Each T5 adapter applies LayerNorm and a linear map to normalized neural features scaled by 32. Learned position embeddings locate memory vectors relative to the target event. Candidate token sequences form a prefix tree: the decoder normalizes logits over the allowed next tokens at each branch and sums log-probabilities along each candidate's path to obtain $\ell_{ik}$. We train the adapters with class-weighted negative log-likelihood using weights $N/(Kn_k)$, where $N$ is the number of training events. Training uses seed~42, batch size 256, AdamW with learning rate $10^{-3}$ and matrix weight decay $10^{-3}$, gradient clipping at 5.0, and a 30-epoch cosine schedule ending at $10^{-5}$, for 36,870 updates. At saved epochs 1, 2, 5, 10, 20 and 30, we fit $\alpha$ on training data by class-weighted cross-entropy and evaluate the fused validation mean. This selects epoch~10 for group memory and epoch~20 for target-only memory. The target-only branch scores 62.55\% alone and 64.22\% after fusion on validation; group memory scores 19.18\% and 64.23\%, respectively.

\section{Implementation Control and Configuration Audit}
\label{app:control}

To measure the two repairs jointly, we run the unmodified official source at commit \path{97da041} with the same V2 initialization, public training data, fixed-50 Subject~0 validation criterion, seed~42 and two-epoch budget (31,226 updates) as checkpoint~A. This preserves evaluation-time attention dropout and the omission of loss temperature and bias from optimization and checkpoint saving. The control scores 57.53\%, 51.51\%, 59.48\% and 43.65\% on the four corpora, for a mean of 53.04\%. The corrected implementation reaches 60.21\%, establishing the combined 7.17-point gain under this configuration. The control predictions were exported with evaluation seed~42 and the original attention dropout enabled. The \href{https://github.com/OpenUploading/RUNE/tree/main/experiments/official_control_20261122}{control archive} includes the configuration, source verification, predictions and results.

The configuration audit distinguishes the published description, released defaults and settings used in our runs. Released defaults allow 50 epochs, select on a frequency-based vocabulary of 250 words, and include all configured subjects in validation. Our matched comparison selects within two epochs on fixed-50 Subject~0 validation. The public V2 initialization has 113,000 pretraining steps on five datasets, compared with 35,000 steps on three datasets for the older public checkpoint. The loader also halves already-reduced recordings from the broader subject pool, retaining approximately the first quarter of the original sessions for Subjects 13--22 and the first eighth for Subjects 23--32. Both sides of the repair comparison use the same public files, initialization and selection rule.

The \href{https://github.com/OpenUploading/RUNE/blob/main/docs/fusion.md}{archived fusion replay} matches all 3,172 fixed-vocabulary test events and candidate orders. For target-only T5, seven recovered hits contribute 0.291286 points and three lost hits subtract 0.147129, giving 0.144157 points. Group-memory fusion recovers one hit and loses one, for 0.019248 points.

\section{Metric, Bootstrap and Per-Class Analysis}
\label{app:reproduction}

For corpus $c$, let $\mathcal{Y}_c$ be the classes present in its test partition and $I_{ck}$ the indices of its events of class $k$. Then
\begin{equation}
  \mathrm{BA}_{10,c}=\frac{1}{|\mathcal{Y}_c|}\sum_{k\in\mathcal{Y}_c}\frac{1}{|I_{ck}|}\sum_{i\in I_{ck}}\mathbf{1}\{k\in\operatorname{Top10}(s_i)\},
  \qquad
  M=\frac{1}{4}\sum_c\mathrm{BA}_{10,c}.
\end{equation}
Ranking always uses all fifty candidates. The test partitions contain 1,364 (Sherlock), 491 (TIMIT), 518 (MOCHA-TIMIT) and 799 (TheMoth) fixed-vocabulary events, with 45, 48, 48 and 47 classes present.

\paragraph{Partitions.}
For Sherlock, the second half of Subject~0 session~11 is used for validation and session~12 for test; the first half of session~11 is in training. For TheMoth, sessions~29 and~30 are used for validation and test. TIMIT follows the official speaker partitions; training speakers who share a sentence with a test speaker are removed, and the two dialect sentences read by every speaker are excluded from validation and test. For MOCHA-TIMIT, sessions~1 and~4 are used for training, the first half of the distinct stimuli in session~2 for validation, and the session-3 stimuli not assigned to validation for test.

\paragraph{Bootstrap.}
We identify each sentence by corpus, session and sentence index; the four test partitions contain 166, 175, 218 and 238 sentences. Using seed~20261118, each bootstrap replicate samples sentences with replacement within each corpus, recomputes per-class hit rates with the sampled multiplicities, and averages the four corpus scores. Classes absent from a replicate are omitted from that corpus's average. The resulting distributions are shifted relative to the observed estimates, upward for the readout differences. We report the basic interval $[2\hat\theta-q_{97.5},\,2\hat\theta-q_{2.5}]$, which reflects the percentile bounds about the observed estimate $\hat\theta$. For comparison, the unreflected percentile intervals are $[+1.10,+4.71]$ for cross-entropy versus squared loss, $[+2.28,+6.99]$ for cross-entropy versus checkpoint-B similarity scoring, and $[-5.24,-1.01]$ for standardized versus raw similarity scores. Both interval forms exclude zero for these three contrasts. Word groups may cross sentence boundaries; Sherlock, MOCHA-TIMIT and TheMoth each contribute one test recording session, while TIMIT contributes 14.

\paragraph{Per-class analysis.}
Nomination rates, per-class hit rates and AUROC describe the pooled test events. Contributions to the primary metric instead preserve the corpus weighting: class $k$ contributes $\tfrac14\sum_c \Delta_{ck}/|\mathcal{Y}_c|$, summing over corpora where it is present, with $\Delta_{ck}$ the change in its hit rate in corpus $c$. These terms sum exactly to the change in the four-corpus mean. Because pooled hit rates weight events rather than corpora, equal pooled rates in Table~\ref{tab:perclass} can coexist with a nonzero contribution to the primary metric.

Comparing checkpoint-B similarity scoring with checkpoint-A cross-entropy, 837 previously missed events become hits and 186 former hits are lost, for a net recovery of 651 events, approximately 21\% of the pooled test set. The six zero-hit classes account for 734 recovered events. Among the other 44 words, 10 contribute positively to the four-corpus change, 26 negatively and 8 contribute zero. Their mean AUROC changes from 0.816 to 0.812, with individual changes of up to 0.15 in either direction. These counts describe event recovery separately from the corpus-balanced metric decomposition.

\paragraph{Corrections of the similarity scores.}
The corrections operate on similarity scores $s_{ik}=x_i^{\top}\hat e_k$ for the 3,172 test events while retaining the text embeddings as weights. Standardization subtracts each candidate's test-event mean and divides by its standard deviation. CSLS uses $2s_{ik}-r_i-r_k$, where $r_i$ averages an event's $m$ largest candidate scores (all fifty for $m\ge50$) and $r_k$ averages a candidate's $m$ largest event scores. We evaluate $m\in\{5,10,50,100\}$. Inverted softmax uses $\exp(\beta s_{ik})/\sum_j\exp(\beta s_{jk})$, with nine values of $\beta$ from 10 to 400. Log-prior adjustment uses $\tau s_{ik}+\gamma\log(n_k/N)$, with $\tau\in\{20,50,80,120,200\}$ and $\gamma\in\{0.25,0.5,1\}$, producing 13 distinct rankings. Table~\ref{tab:diagnostics} reports the settings with the highest test mean for these three searches: $m=100$, $\beta=200$, and $(\tau,\gamma)=(50,0.25)$.

For the fitted temperature and biases, we minimize cross-entropy on $\tau s_{ik}+b_k$, weighting events by $1/n_k$ and penalizing biases by $10^{-3}\sum_k b_k^2$. We split test sentences in half within each corpus, fit on one half and score the other, then reverse the roles. Averaging over 20 random splits gives the reported cross-fitted result, with a standard deviation of 0.54 points. Varying the penalty from 0 to 1 and using either the original weights or weights balancing classes within the fitting half gives means from 56.06\% to 60.00\%.

Metric-maximizing offsets use greedy coordinate ascent over a grid of offsets for each candidate, directly optimizing the four-corpus mean. Starting from zero and cross-fitting over five sentence splits gives a standard deviation of 1.07 points. Fitting and scoring on the complete test set provides an in-sample reference: the search reaches 64.31\% from zero offsets, leaving \emph{the}, \emph{of} and \emph{i} at zero hit rate. Three other starting points yield 64.01--64.26\%; initialization by negative column means gives every previously zero-hit class a positive hit rate. Table~\ref{tab:diagnostics} reports the zero-initialized run.

\paragraph{Artifacts.}
The repository, \url{https://github.com/OpenUploading/RUNE}, holds source code, configurations, result records and the per-event test scores analysed above. Raw recordings, feature caches and model checkpoints are obtained separately. The data snapshot is revision \path{78443d024480c66bf340073c69bef50af8d1d335} of \path{pnpl/LibriBrain2}, and the per-corpus results for all 20 development variants are included in the release. The scripts in \path{tools/} recompute every reported mean (\path{recompute_corpus_means.py}), the intervals (\path{bootstrap_readout_intervals.py}), and the per-class diagnostics (\path{readout_diagnostics.py}).

\begin{table}[h]
  \centering
  \scriptsize
  \caption{Per-class results on the pooled test set, classes ordered by number of training events. Hit rate: Top-10 hit rate (\%). AUROC: one-versus-rest area under the ROC curve of the class's score. ``sim.'' is similarity scoring on checkpoint~B and ``CE'' the cross-entropy readout on checkpoint~A. The candidate list is lower-cased, so \emph{i} is the pronoun.}
  \label{tab:perclass}
  \begin{tabular}{lrrrrrr@{\hspace{2.2em}}lrrrrrr}
\toprule
Word & Train & Test & \multicolumn{2}{c}{Hit rate} & \multicolumn{2}{c}{AUROC} & Word & Train & Test & \multicolumn{2}{c}{Hit rate} & \multicolumn{2}{c}{AUROC} \\
 & & & sim. & CE & sim. & CE &  & & & sim. & CE & sim. & CE \\
\midrule
\emph{the} & 41,876 & 486 & 0 & 83 & 0.40 & 0.86 & \emph{all} & 3,353 & 29 & 41 & 66 & 0.69 & 0.74 \\
\emph{and} & 21,613 & 316 & 79 & 80 & 0.84 & 0.83 & \emph{on} & 3,343 & 62 & 71 & 65 & 0.77 & 0.78 \\
\emph{of} & 18,985 & 173 & 0 & 78 & 0.54 & 0.84 & \emph{so} & 3,116 & 25 & 68 & 68 & 0.80 & 0.86 \\
\emph{i} & 18,901 & 94 & 0 & 70 & 0.56 & 0.76 & \emph{an} & 3,016 & 16 & 94 & 88 & 0.80 & 0.84 \\
\emph{to} & 18,663 & 176 & 78 & 74 & 0.79 & 0.76 & \emph{she} & 2,904 & 59 & 0 & 68 & 0.72 & 0.80 \\
\emph{a} & 18,101 & 213 & 96 & 93 & 0.94 & 0.94 & \emph{were} & 2,685 & 33 & 61 & 52 & 0.68 & 0.66 \\
\emph{in} & 13,111 & 127 & 61 & 59 & 0.73 & 0.75 & \emph{are} & 2,569 & 50 & 74 & 78 & 0.79 & 0.80 \\
\emph{that} & 12,776 & 77 & 48 & 44 & 0.74 & 0.71 & \emph{out} & 2,181 & 35 & 83 & 83 & 0.83 & 0.88 \\
\emph{it} & 11,626 & 58 & 72 & 50 & 0.69 & 0.73 & \emph{very} & 2,092 & 16 & 88 & 75 & 0.96 & 0.95 \\
\emph{he} & 11,487 & 153 & 70 & 61 & 0.76 & 0.76 & \emph{will} & 2,078 & 10 & 50 & 20 & 0.68 & 0.53 \\
\emph{was} & 11,419 & 118 & 69 & 64 & 0.84 & 0.84 & \emph{they} & 2,027 & 47 & 62 & 64 & 0.79 & 0.73 \\
\emph{is} & 7,715 & 53 & 64 & 58 & 0.79 & 0.73 & \emph{do} & 1,934 & 12 & 67 & 33 & 0.79 & 0.69 \\
\emph{had} & 6,953 & 80 & 66 & 64 & 0.81 & 0.79 & \emph{our} & 1,894 & 14 & 79 & 79 & 0.78 & 0.78 \\
\emph{my} & 6,110 & 27 & 59 & 59 & 0.80 & 0.77 & \emph{has} & 1,805 & 14 & 43 & 21 & 0.66 & 0.55 \\
\emph{have} & 6,002 & 37 & 51 & 41 & 0.76 & 0.69 & \emph{can} & 1,597 & 16 & 69 & 69 & 0.86 & 0.86 \\
\emph{for} & 5,443 & 82 & 0 & 70 & 0.63 & 0.74 & \emph{time} & 1,201 & 15 & 73 & 73 & 0.95 & 0.97 \\
\emph{as} & 5,129 & 47 & 79 & 64 & 0.82 & 0.77 & \emph{am} & 1,112 & 3 & 67 & 67 & 0.70 & 0.85 \\
\emph{at} & 5,056 & 41 & 78 & 59 & 0.76 & 0.72 & \emph{any} & 1,011 & 10 & 90 & 90 & 0.93 & 0.94 \\
\emph{we} & 4,832 & 49 & 0 & 71 & 0.75 & 0.70 & \emph{think} & 965 & 9 & 67 & 89 & 0.95 & 0.95 \\
\emph{but} & 4,771 & 39 & 79 & 79 & 0.87 & 0.87 & \emph{good} & 915 & 5 & 60 & 20 & 0.76 & 0.75 \\
\emph{this} & 4,489 & 43 & 84 & 86 & 0.84 & 0.87 & \emph{it's} & 660 & 16 & 44 & 94 & 0.75 & 0.84 \\
\emph{not} & 4,119 & 33 & 70 & 64 & 0.89 & 0.89 & \emph{really} & 455 & 7 & 100 & 100 & 0.98 & 0.99 \\
\emph{be} & 4,033 & 47 & 70 & 60 & 0.78 & 0.74 & \emph{always} & 374 & 11 & 82 & 82 & 0.97 & 0.98 \\
\emph{there} & 3,629 & 38 & 76 & 71 & 0.78 & 0.82 & \emph{new} & 339 & 7 & 100 & 100 & 0.97 & 0.96 \\
\emph{him} & 3,596 & 36 & 81 & 75 & 0.85 & 0.89 & \emph{people} & 310 & 8 & 88 & 100 & 0.96 & 0.98 \\
\bottomrule
\end{tabular}

\end{table}

\FloatBarrier

\end{document}